 \documentclass[pmlr,twocolumn,10pt,preprint]{jmlr} 

\usepackage{comment}
\newif\ifanon
\anonfalse   

\ifanon
  \includecomment{anon}   
  \excludecomment{camera} 
\else
  \includecomment{camera}
  \excludecomment{anon}
\fi

\ifanon
  \usepackage{hyperref}
  \hypersetup{pdfauthor={Anonymous}, pdftitle={}}
\fi

\usepackage{booktabs}
\usepackage{siunitx}

\usepackage[switch]{lineno}

\theorembodyfont{\upshape}
\theoremheaderfont{\scshape}
\theorempostheader{:}
\theoremsep{\newline}

 \title[In-Domain Supervised Pathology Report Classification]{In-Domain Supervised Pathology Report Classification: A Reproducible Pipeline from Data Curation to Production-Matched Evaluation}

\begin{camera}

\author{
    \Name{Isaac Hands} \Email{isaac.hands@uky.edu}\\
    \addr Department of Computer Science, University of Kentucky, Lexington, KY, USA\\
    \addr UK Markey Cancer Center, Lexington, KY, USA\\
    \addr Kentucky Cancer Registry, Lexington, KY, USA
\AND
    \Name{Bin Huang} \Email{bhuan0@uky.edu}\\
    \addr Division of Cancer Biostatistics, University of Kentucky, Lexington, KY, USA\\
    \addr UK Markey Cancer Center, Lexington, KY, USA\\
    \addr Kentucky Cancer Registry, Lexington, KY, USA
\AND
    \Name{Adam Spannaus} \Email{spannausat@ornl.gov}\\
    \addr Oak Ridge National Laboratory, Oak Ridge, TN, USA
\AND
    \Name{John Gounley} \Email{gounleyjp@ornl.gov}\\
    \addr Oak Ridge National Laboratory, Oak Ridge, TN, USA
\AND
    \Name{Heidi Hanson} \Email{hansonha@ornl.gov}\\
    \addr Oak Ridge National Laboratory, Oak Ridge, TN, USA
\AND
    \Name{Eric Durbin} \Email{ericd@kcr.uky.edu}\\
    \addr Division of Biomedical Informatics, University of Kentucky, Lexington, KY, USA\\
    \addr UK Markey Cancer Center, Lexington, KY, USA\\
    \addr Kentucky Cancer Registry, Lexington, KY, USA
\AND
    \Name{Sally R. Ellingson} \Email{sel228@uky.edu}\\
    \addr Division of Biomedical Informatics, University of Kentucky, Lexington, KY, USA\\
    \addr UK Markey Cancer Center, Lexington, KY, USA
}

\end{camera}

\begin{document}

\maketitle
\thispagestyle{plain} 
\pagestyle{plain}     

\begin{abstract}

We introduce an in-domain supervised pipeline designed to counter the out-of-distribution performance drop that hampers supervised biomedical NLP models, a problem observed when models trained on pathology reports are moved across cancer registries\citep{rios2019cross}. Our contribution is a reproducible recipe for training a supervised classifier from routinely collected cancer registry data. It describes how to build the in-domain training set and a production-matched holdout, and to choose operating points that keep the false-negative rate (FNR) very low while keeping reviewer workload manageable. The pipeline standardizes data curation with facility-stratified sampling and separate handling of reports linked to registry cases, and includes a blinded manual audit to estimate positive-case prevalence and label noise. On a 418k-report holdout set, the Kentucky model achieved FNR 0.003 and false-positive rate (FPR) 0.097, improving over the Seattle-trained MOSSAIC OncoID baseline (FNR 0.010, FPR 0.183) and raising F1 from 0.860 to 0.922. In a blinded manual review of 600 reports, estimated positive prevalence declined from 0.500 to 0.398, indicating substantial label noise with errors concentrated in rare primary sites.

\end{abstract}
\begin{keywords}
cancer registry, natural language processing, pathology reports, deep learning, neural networks, clinical NLP, distribution shift, in-domain training, threshold selection
\end{keywords}

\paragraph*{Data and Code Availability}
Our data consisted of pathology reports from human patients which included protected health information in both the structured and unstructured portions of the documents. These data were received at the SEER Kentucky Cancer Registry as part of state-mandated cancer reporting and none of the documents are available to other researchers due to privacy and confidentiality regulations. 
\begin{camera}
The software used to build our classification models has been released previously as part of the NCI MOSSAIC project \citep{fresco, bardi}.
\end{camera}
\begin{anon}
The software used to build our classification models has been released as ANONYMIZED.
\end{anon}

\paragraph*{Institutional Review Board (IRB)}
\begin{camera}
This study received approval as IRB Project DOE000152    
\end{camera}
\begin{anon}
This study received IRB approval
\end{anon}

\section{Introduction}
\label{sec:intro}

Cancer is a leading cause of death and disease in the United States, with Kentucky reporting the highest incidence and the second-highest mortality in recent years \citep{statecancerprofiles}. Counting and tracking cancer incidence and mortality rates in the state of Kentucky is the primary responsibility of the SEER Kentucky Cancer Registry (KCR). For many cancers, the first definitive diagnosis appears in the narrative text of a pathology report, which is frequently reviewed manually in a time-consuming, high-volume task at KCR. The MOSSAIC OncoID deep learning classifier was developed to automatically identify reportable cancer terminology in pathology reports, but performance can vary across registries due to heterogeneous report styles, sources, and workflows.

In this work, we demonstrate improved performance of the OncoID classifier by training a new model from scratch with in-domain supervised learning on target-domain data curated entirely from KCR’s routine pathology report feeds. To standardize curation, we sample by facility and separately handle reports linked to registry cases versus unlinked reports. Before training, we exclude likely but unvalidated negatives to reduce label noise and avoid biasing operating-point selection. Evaluation uses a production-matched holdout that mirrors what staff see in deployment. We then choose operating points that prioritize very low false-negative rates while keeping false positives at a level reviewers can manage, and we include a blinded manual audit to estimate positive-case prevalence and label noise.

We aim to share a reproducible data-to-deployment strategy that other cancer registries can adopt to build pathology report classifiers, evaluate under production-matched conditions, and select operating points that maintain case capture while keeping review volume manageable for staff.

\section{Related Work}
NLP methods for classifying and extracting data from pathology reports have been explored for at least the last 25 years and summarized in systematic reviews\citep{burger2016natural, santos2022automatic}.  Many of these efforts utilized supervised machine learning techniques where manually labeled training data was required. These curated datasets are difficult to assemble because they require expert medical knowledge, contain private health information, and often do not follow common data standards across healthcare systems. As a result, many supervised NLP models for pathology reports were trained on datasets that represent a single cancer type such as prostate\citep{khosravi2021deep} or breast\citep{saib2020hierarchical}, or consist of only a small sample of reports from a single facility \citep{schroeck2017development, martinez2011information, yoon2018filter}. Recognizing the difficulty of assembling cancer pathology datasets that are representative of all cancers, the SEER program created the Modeling Outcomes using Surveillance data and Scalable Artificial Intelligence for Cancer (MOSSAIC) project\citep{hsu2024machine}. This unique collaboration across federal government organizations and state-based cancer registries develops advanced NLP tools using, ``a rich source of high-volume annotated data'', from 6 SEER population-based cancer registries including KCR. 

MOSSAIC developed a classifier, OncoID, based on the Hierarchical Self-Attention Network architecture (HiSAN) \citep{gao2019classifying}, to distinguish pathology reports as either \textit{reportable} or \textit{non-reportable} to SEER cancer registries. HiSAN outperformed traditional approaches such as Naïve Bayes, Logistic Regression, CNNs, and other neural network models. The primary goal of OncoID was to reduce manual effort at cancer registries without missing cancer cases by accurately identifying reports containing SEER-reportable findings and filtering out all others. The initial OncoID model was trained on pathology reports from the Seattle SEER registry, where complete sets of positive and negative reports were readily available. While the model achieved high accuracy (98.6\%) on Seattle data, its deployment at KCR resulted in a high false positive rate, substantially increasing the manual review burden for \textit{non-reportable} reports. In this work, we demonstrate that a Kentucky-trained OncoID model significantly improves performance and has strong potential to reduce manual review workload at KCR.

\begin{figure*}[h]
 \centering
    \includegraphics[width=\textwidth]{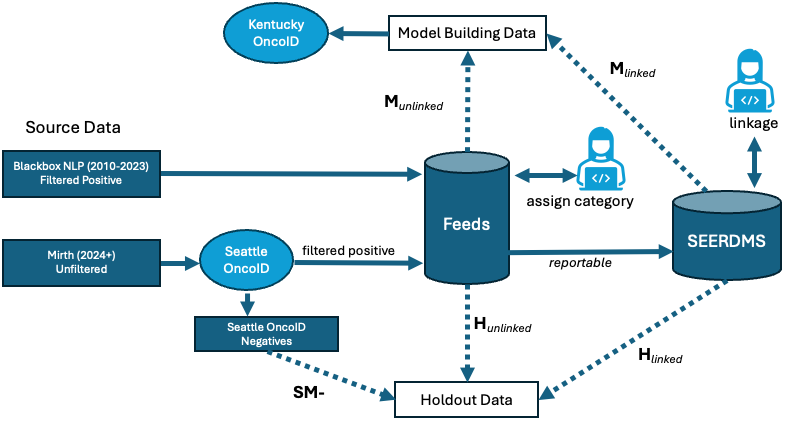} 
    \caption{Overview of Processing Procedures for Source Data and Dataset Creation}
    \label{fig:data}
\end{figure*}

\section{Methods}

\subsection{Data}

The source data in Figure \ref{fig:data} indicates all the data we have access to. Prior to November 2023, KCR received data that was prefiltered by a third-party NLP algorithm based on a manually curated set of inclusion and exclusion criteria. This algorithm was a blackbox filter such that pathology reports labeled as \textit{non-reportable} were never delivered to the cancer registry. From November 2023 until present, KCR has been receiving unfiltered pathology reports from a majority of reporting facilities in the state. This data is filtered by a MOSSAIC model trained on data from the Seattle Cancer Registry, which we will call the Seattle OncoID model, or $SM$. Everything that is labeled \textit{reportable} through the $SM$ filter and all data filtered by the blackbox NLP algorithm are transferred into the KCR Feeds database. From there, the pathology reports are categorized by a single Oncology Data Specialist (ODS) in a rapid manner that prioritizes retention of all possible reportable cancer cases. The categories used during this brief manual review are: Needs Review, History of Cancer, Recurrence, Metastases, No Cancer, and a few project specific labels that identify \textit{non-reportable} cancer conditions important to the registry. In the Feeds database, pathology reports are utilized for casefinding and case abstraction at KCR and hospitals across the state. In addition, all pathology reports from Feeds are sent to the KCR central database software, called SEERDMS, where another ODS links the reports with consolidated patient records when available.

We built the dataset for this project by collecting three subsets of our source data so that we can stratify by different variables and ensure we have representation across our model building and holdout data. The three subsets of pathology reports are: (1) reports linked to cancer abstracts in SEERDMS, (2) reports not linked to abstracts in SEERDMS, and (3) reports that $SM$ labeled as \textit{non-reportable}. It is important to note that the proportion of \textit{reportable} to \textit{non-reportable} reports received at KCR is not stable over time. This imbalance occurs because the black-box NLP filter, which only allowed filtered reports through, was replaced with $SM$ in November 2023. In addition, reporting facilities are added and removed from the KCR reporting system over time, and they each have different proportions of \textit{reportable} to \textit{non-reportable} reports due to their varying patient populations.

First, we took all data from the Feeds database that had an assigned category and split it by whether it had been linked to cases in SEERDMS or not. The categorized reports in the Feeds database have gone through either the blackbox NLP filter or $SM$, and then were rapidly reviewed by an ODS, so we expect their labels to be more accurate than reports that have only been filtered automatically. The reports that are linked in SEERDMS were then further stratified by the topography of the linked case in order to ensure broad representation across cancer sites and split 80:20 for model building ($M_{linked}$) and holdout ($H_{linked}$). The data not linked to cases in SEERDMS does not have any topography information for stratification, but does have one of the following categories assigned by an ODS in the Feeds database:  Needs Review, Metastases, Recurrence, History of Cancer, No Cancer, and Project Specific. This unlinked data is stratified by category and split 90:10 for model building ($M_{unlinked}$) and holdout ($H_{unlinked}$). For these reports, we converted the ODS assigned categories into binary classes of \textit{reportable} (Needs Review, Metastases, Recurrence, History of Cancer) and \textit{non-reportable} (No Cancer, Project Specific). By utilizing a different split percentage between linked (80/20) and unlinked (90/10) data sets, we were able to more closely match the observed proportion of linked to unlinked data in our production data streams.




Preliminary experiments showed better performance when training on pathology reports categorized by an ODS after filtering by either the blackbox NLP model or $SM$. Accordingly, we excluded $SM$-labeled \textit{non-reportable} reports ($SM_-$) from training because their reportability has not been human-validated. We included them in the holdout set to reflect the real registry data stream by adding a facility-stratified random sample of $SM_-$ roughly equal in size to $H_{linked}+H_{unlinked}$.

\subsection{Model}

\begin{camera}
The MOSSAIC classifier is a hierarchical self-attention network model created from two software packages, BARDI (Batch-processing Abstraction for Raw Data Integration) \citep{bardi} and FrESCO (Framework for Exploring Scalable Computational Oncology) \citep{fresco}. Vocabulary building and file preprocessing was performed with BARDI on a 32-core Intel(R) Xeon(R) Gold 6326 CPU @ 2.90GHz. The model building and inference was performed with FrESCO on dual NVIDIA H100 GPUs running CUDA 12.7.
\end{camera}
\begin{anon}
The MOSSAIC classifier is a hierarchical self-attention network model created from two software packages, ANONYMIZED and ANONYMIZED. Vocabulary building and file preprocessing was performed on a 32-core Intel(R) Xeon(R) Gold 6326 CPU @ 2.90GHz. The model building and inference was performed on dual NVIDIA H100 GPUs running CUDA 12.7.
\end{anon}

During the modeling phase, the model data, $M$, was randomly split into 70\% training, 15\% validation, and 15\% testing. The vocabulary was built using this training split, a minimum word count of 5, a regular expression to remove special characters, and an exclusion list that removed some common first names. Due to the class imbalance in the data and the operational requirement to prioritize recall, we tried several weights for the loss function of the model including 30 to 1, 15 to 1, and none, as the \textit{non-reportable} to \textit{reportable} weights.




\subsection{Manual Review}

We anticipated some amount of label noise in our data set, which we sought to characterize, along with a detailed uncertainty quantification, through a two-stage manual review process. The first stage was a systematic manual review of SEER reportability of the pathology reports while the second stage was a qualitative review of misclassified reports. Our dataset labels are assigned by a single ODS in a rapid manner with a priority on recall (not missing any cases) and high throughput. During our systematic manual review, two ODS staff not involved in the original category assignment worked independently to assign a class of either \textit{reportable} or \textit{non-reportable} to a small sample of the holdout set. Once the sample dataset was reviewed and classified by the first two reviewers, a third independent reviewer adjudicated any conflicting class assignments and assigned a final class to each report. Each reviewer was instructed to read the pathology report in its entirety and assign a reportability class based solely on the text in the single report according to the 2024 SEER Coding and Reportability Guidelines \citep{seercoding2024}.

\subsubsection{Sample Size for Manual Review}

The sample size for the systematic manual review was 600 total pathology reports, 300 from the \textit{reportable} class and 300 from the \textit{non-reportable} class, drawn randomly from the overall holdout set labels. From this first stage of manual review, we estimated the true positive prevalence rate of the pathology reports in our holdout set. Assuming the null hypothesis of 50\% positive prevalence rate and a statistical significance level of 0.05, the sample size of 600 will have 0.8381 power to detect a statistical significance if the true positive prevalence rate is 0.44, or 0.9987 power if the true positive prevalence rate is 0.40, based on a two-sided, one-sample exact test\citep{fleiss2013statistical}.

\subsubsection{Qualitative Review of Pathology Reports}

The second stage of the manual review process was for the authors to read through a small sample of the pathology reports that were misclassified by the KY OncoID classifier and try to identify patterns in the predictions. This sample of reports was taken by first identifying the false negative reports from the holdout set, those assigned the \textit{reportable} class in the dataset but predicted to be \textit{non-reportable} by the KY OncoID model. Likewise, the false positive reports were identified, those assigned the \textit{non-reportable} class but predicted to be \textit{reportable}. From these two subsets of the holdout predictions, 30 reports each were sampled, reviewed by the authors, and qualitatively assessed for patterns in the reported findings.

\section{Results and Discussion}

\subsection{Model Building} The total data used for model building, $M_{linked} + M_{unlinked}$, contained 904,291 \textit{reportable} documents and 476,242 \textit{non-reportable} documents for a total of 1,380,533 pathology reports and a \textit{reportable} to \textit{non-reportable} ratio of 1.9. There was a minimum, maximum, and average report word count of 9, 22,736, and 434 respectively and 2,297 reports had more than 3,000 words (a cut-off used in the embedding). The resulting vocabulary built from the training data contained 117,751 words. Our final KY OncoID model trained for 15 epochs with balanced metrics across the train, validation, and test splits during the modeling phase. 





\subsection{Holdout Data}

We kept a complete holdout set, $H$, to assess performance across different variations of modeling with a consistent set that was stratified by important features as described in the Methods section. The total data in the holdout set, $H_{linked} + H_{unlinked} + SM_-$, contained 154,580 \textit{reportable} documents and 263,586 \textit{non-reportable} documents for a total of 418,166 pathology reports and a \textit{reportable} to \textit{non-reportable} ratio of 0.586. There was a minimum, maximum, and average report word count of 10, 22,774, and 351 respectively, and 434 reports that had more than 3,000 words.

The F1-score, precision, recall, false positive rate (FPR), and false negative rate (FNR) for different weighting schemes of the loss function are given in Table \ref{tab:loss2} for the complete holdout set, $H$. Our operating-point criteria was met by selecting the model with the lowest FNR while maintaining a FPR of less than 10\% on a holdout dataset that resembles what is seen in production. Using no weighting scheme gave the best F1-score due to the low FPR of 0.021. However, we can tolerate a higher FPR in our manual review in exchange for a lower FNR and optimize to not miss \textit{reportable} cancer documents. A weighting scheme of 30 to 1 gave the best recall due to the low FNR of 0.002; however, the associated FPR is above our desired threshold. Therefore, we chose the 15 to 1 weighting scheme as the KY OncoID model, with FPR of 0.097 and FNR of 0.003. The remaining results will refer only to this model. To better understand how our model performs on different data, we present a complete breakdown of metrics by holdout set type in Table \ref{tab:holdout}.





\begin{table}[t]
\floatconts
  {tab:loss2}
  {\caption{Metrics by Loss Function Weighting Scheme}}
  {%
    \scriptsize
    \setlength{\tabcolsep}{4pt} 
    \begin{tabular*}{\linewidth}{@{\extracolsep{\fill}} l c c c c c @{}}
      \toprule
      \textbf{Weights} & \textbf{F1} & \textbf{Prec.} & \textbf{Rec.} & \textbf{FPR} & \textbf{FNR} \\
      \midrule
      none   & \textbf{0.971} & \textbf{0.965} & 0.976 & \textbf{0.021} & 0.024 \\
      15:1   & 0.922 & 0.858 & 0.997 & 0.097* & 0.003 \\
      30:1   & 0.901 & 0.821 & \textbf{0.998} & 0.128 & \textbf{0.002} \\
      \bottomrule
    \end{tabular*}
  }
\end{table}

\begin{table}[t]
\floatconts
  {tab:holdout}
  {\caption{Metrics by Holdout Set}}
  {%
    \scriptsize
    \setlength{\tabcolsep}{4pt}
    \begin{tabular*}{\linewidth}{@{\extracolsep{\fill}} l c c c c c @{}}
      \toprule
      \textbf{Set} & \textbf{F1} & \textbf{Prec.} & \textbf{Rec.} & \textbf{FPR} & \textbf{FNR} \\
      \midrule
      $H_{total}$       & 0.922 & 0.858 & 0.997 & 0.097 & 0.003 \\
      $H_{linked}$      & \textbf{0.989} & \textbf{0.979} & \textbf{0.999} & 0.789 & \textbf{0.001} \\
      $H_{unlinked}$ & 0.890 & 0.806 & 0.993 & 0.265 & 0.007 \\
      $SM_{-}$          &  --   &  --   &  --   & \textbf{0.046} &  -- \\
      \bottomrule
    \end{tabular*}
  }
\end{table}


\subsubsection{Linked Holdout Data}
The $H_{linked}$ set of the holdout data contains reports that are linked to a case in SEERDMS. Although not all of these pathology reports will have reportable findings, they are all associated with patients that have reportable cancer cases, and therefore this subset has a higher proportion of pathology reports with the \textit{reportable} class than the other subsets. All of this data has been reviewed by a human for both categorization and linkage. This subset has the highest FPR of 0.789, due to the relatively low total number of \textit{non-reportable} reports, but the $H_{linked}$ set performs better in precision and recall than the other sets, resulting in the highest F1 score. An unusual finding is that both categories associated with our \textit{non-reportable} class have a higher number of reports that are predicted by the model to be reportable. Roughly twice as many ``No cancer'' reports are classified as \textit{reportable} vs \textit{non-reportable} (n=673) and nearly 7 times as many ``Project-Specific'' reports are classified as \textit{reportable} vs. \textit{non-reportable} (n=1379). A potential explanation for this is that although these reports are linked to a case in SEERDMS, they may not contain reportable findings that led to a new reportable cancer case, but instead simply contain an indication that the patient in the reports was in the process of a diagnosis or subsequent cancer care. 

\begin{figure}
    \centering
    \includegraphics[width=0.95\linewidth]{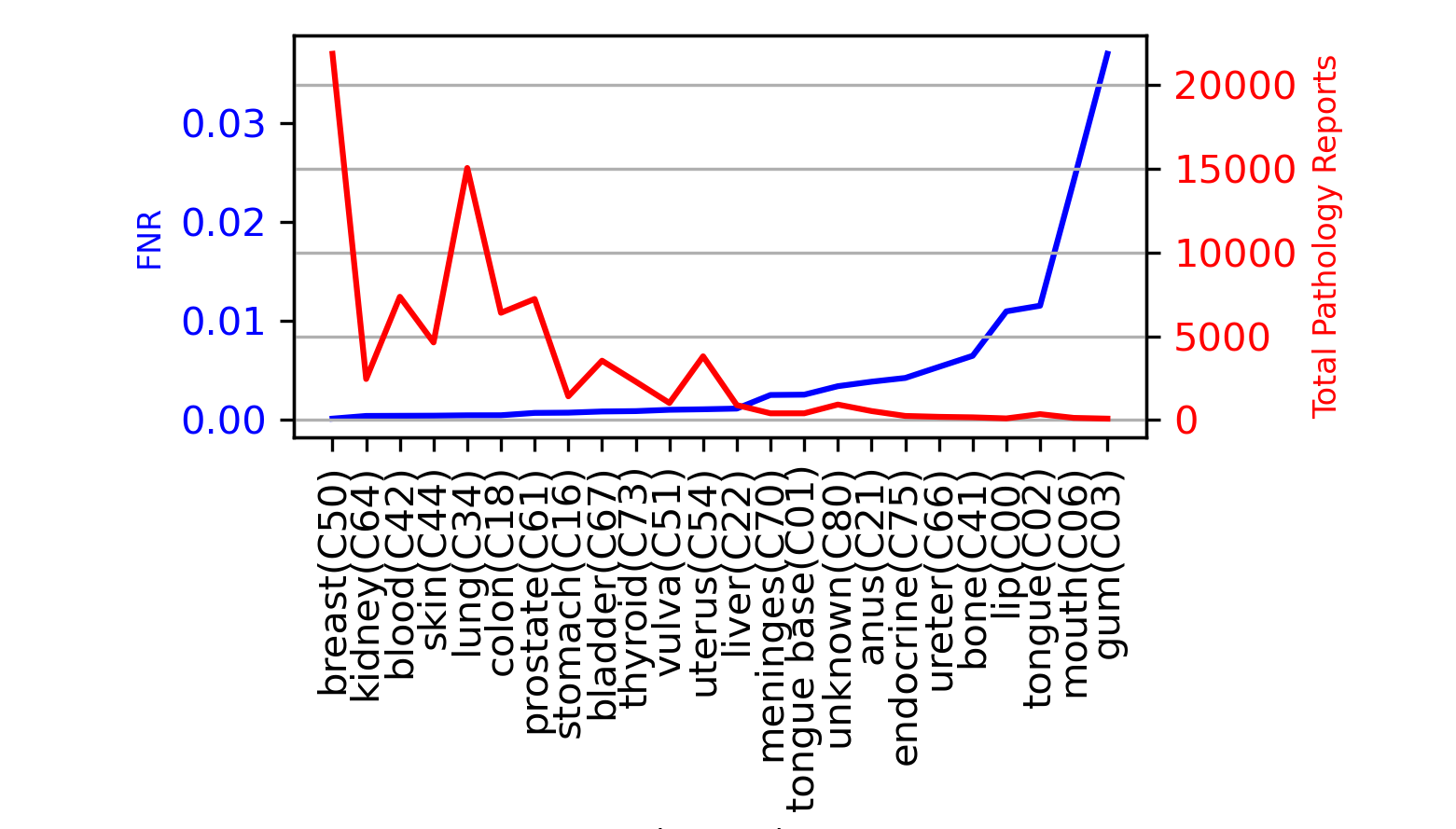}
    \caption{FNR by Primary Site}
    \label{fig:site}
\end{figure}

Since the $H_{linked}$ data is linked to a SEER cancer case, these reports have an associated primary cancer site which was used to stratify the data for our model training and holdout data. We use this information to report the non-zero FNRs by primary site in Figure \ref{fig:site}. There were 45 sites with a FNR of zero and not included here. The highest FNRs all come from cancers which are rare in our dataset, such as tongue, mouth, and gum. The primary sites with the largest occurrences in our dataset, breast and lung, both have very low FNRs of 0.0001 and 0.0005, respectively. These results are unsurprising for a supervised machine learning model, where training on a more balanced dataset typically leads to better performance. However, acquiring labeled examples of pathology reports from rare cancers is inherently difficult, especially when trying to capture a breadth of findings within those cancers and across linguistic variations.



\subsubsection{Unlinked Holdout Data}
The $H_{unlinked}$ set is all data that was filtered through either $SM$ or the blackbox NLP algorithm, has been categorized by an ODS, but was not linked to a case in SEERDMS. Although these data passed through an initial NLP filter, there is a higher proportion of reports that were assigned the \textit{non-reportable} class in this dataset than in the $H_{linked}$ set. The $H_{unlinked}$ set has closer to an equal class split between \textit{reportable} and \textit{non-reportable} than the other holdout datasets. Looking at Table \ref{tab:holdout}, $H_{unlinked}$ still has a higher FPR of 0.265 compared to $H_{total}$, but much lower than the 0.789 FPR of $H_{linked}$. Compared to $H_{linked}$, the predictions for the \textit{non-reportable} class categories are more accurate, which can be expected when $H_{linked}$, by nature of being linked to cancer cases, should have more of a signal for the \textit{reportable} classification as opposed to the reports in $H_{unlinked}$. It is noteworthy that the Project-Specific category is being misclassified at a much higher rate compared to the No Cancer category, which might be explained by the fact that the reports assigned to the Project-Specific category are typically utilized in cancer-related projects whereas the No Cancer category is typically assigned to reports that have no mention of cancer.




\subsubsection{Seattle Mossaic Model Negatives}
\begin{figure}
    \centering
    \includegraphics[width=0.95\linewidth]{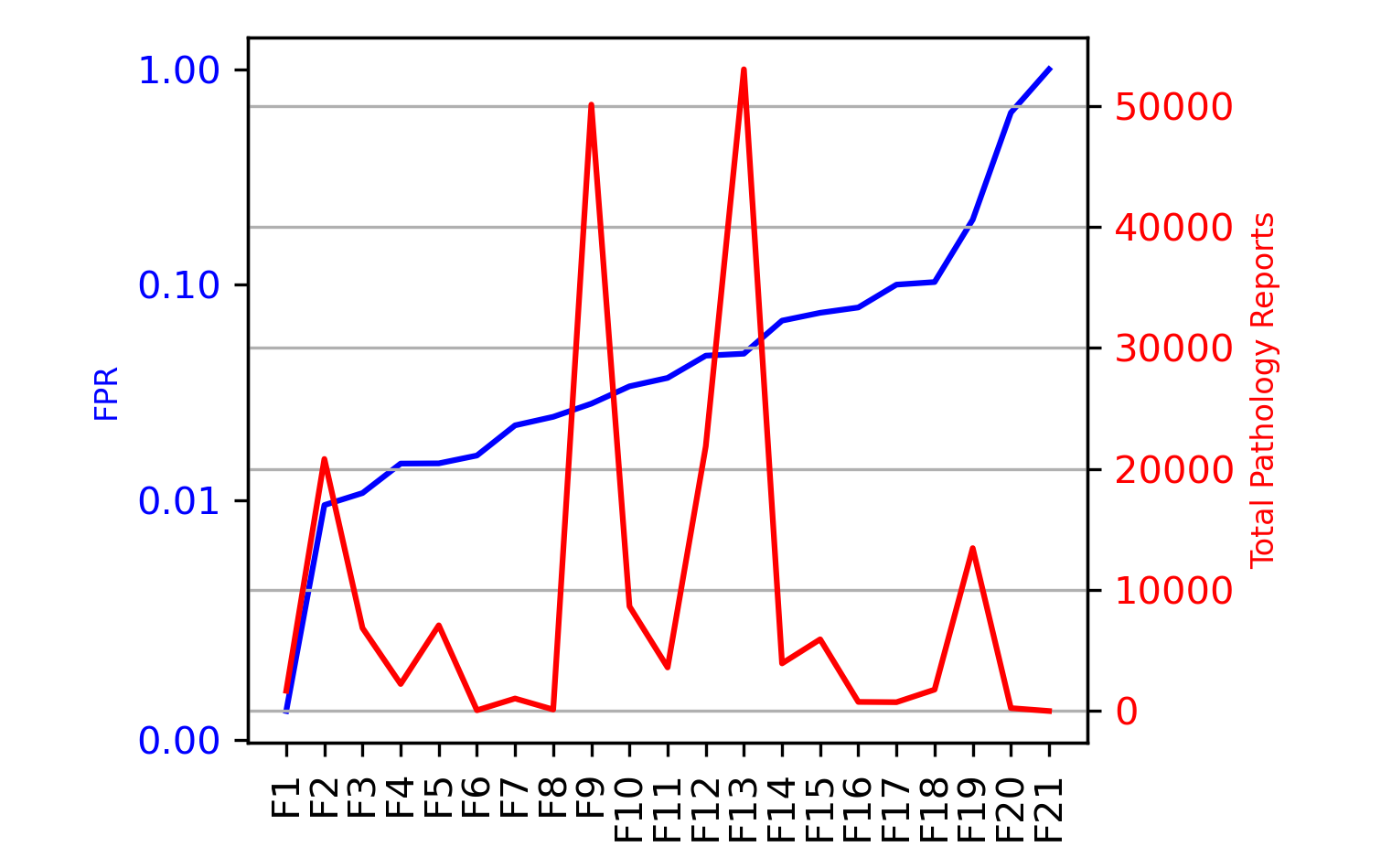}
    \caption{FPR by Reporting Facility}
    \label{fig:facility}
\end{figure}

The $SM_-$ set represents data that was not used to build the model and only used in the holdout set. The reports have not been checked by a human but are all classified as \textit{non-reportable} by $SM$. The only other annotations available for this data are the facility names from which the reports were sent. We report all non-zero FPRs by facility for this subset of data in Figure~\ref{fig:facility}. Note that the FPR is displayed on a logarithmic scale but the real values are listed on the axis. It is evident from looking at this figure that the FPR on reports from a facility does not correlate with the number of pathology reports sent by that facility. A key assumption in building our model from pathology reports in Kentucky rather than relying on a model built from Seattle data is that the diagnostic terminology used in Kentucky may differ from Seattle. It follows that within Kentucky, there may be linguistic variations among pathology laboratories, which could explain the lack of correlation between FPR and number of pathology reports at a facility. Another explanation could be that some pathology facilities within Kentucky, such as F9 and F13 in the figure, diagnose a high proportion of very specific cancers, while our model does not perform uniformly across all cancer types. 

\subsubsection{Comparison to Seattle OncoID Model Performance}
Here we present the results of the holdout data on $SM$ with a 10 to 1 \textit{non-reportable} to \textit{reportable} weighting scheme that is currently in production at KCR. It is worth noting that the KY OncoID model was necessarily trained on data already filtered by $SM$, meaning its performance cannot be viewed as entirely independent. For this reason, we constructed the holdout set to include both $SM$ positives and negatives, providing a more representative comparison. 

We give the metrics by the type of holdout set in Table \ref{tab:SMholdout}. Note that some of the $SM_{-}$ reports are shown as \textit{reportable} even though, by definition, all $SM_{-}$ reports should be \textit{non-reportable}. This discrepancy exists because the parsing logic for converting HL7 to tokenized narrative text has changed over time from when the holdout set was first created to the time of the inference run by the final model. These parsing changes were made due to regular software dependency updates and changes to the data structure sent by pathology labs. As a result, the narrative text parsed out of the same HL7 messages has changed over time, even though $SM$ has not. To compare the two models using only human-labeled reports, the metrics are given for both $SM$ and Kentucky OncoID using $H_{linked}+H_{unlinked}$ in Table \ref{tab:metricshuman}. 

Overall, $SM$ gave an FPR of 0.183 on the complete holdout set, meaning that 18.3\% of all truly \textit{non-reportable} reports in $H$ would have been unnecessarily reviewed by KCR staff compared to 9.7\% using the KY OncoID model. It is even more striking to compare the FPRs of 0.819 versus 0.290 for $SM$ and KY OncoID, respectively, on the subset of $H$ that were labeled by an ODS. This represents an almost 65\% difference in the number of false positive reports that would be reviewed, a significant amount of manual effort.

 
    					
        


\begin{table}[t]
\floatconts
  {tab:SMholdout}
  {\caption{Seattle OncoID Metrics by Holdout Set}}
  {%
    \scriptsize
    \setlength{\tabcolsep}{4pt}
    \begin{tabular*}{\linewidth}{@{\extracolsep{\fill}} l c c c c c @{}}
      \toprule
      \textbf{Set} & \textbf{F1} & \textbf{Prec.} & \textbf{Rec.} & \textbf{FPR} & \textbf{FNR} \\
      \midrule
      $H_{total}$       & 0.860 & 0.760 & 0.990 & 0.183 & 0.010 \\
      $H_{linked}$      & \textbf{0.986} & \textbf{0.977} & \textbf{0.996} & 0.884 & \textbf{0.004} \\
      $H_{unlinked}$ & 0.721 & 0.570 & 0.980 & 0.816 & 0.020 \\
      $SM_{-}$          & --   & --   & --   & \textbf{0.018} & -- \\
      \bottomrule
    \end{tabular*}
  }
\end{table}



			
    								
        


\begin{table}[t]
\floatconts
  {tab:metricshuman}
  {\caption{Metrics on Human Labeled Reports ($H_{linked}+H_{unlinked}$)}}
  {%
    \scriptsize
    \setlength{\tabcolsep}{4pt}
    \begin{tabular*}{\linewidth}{@{\extracolsep{\fill}} l c c c c c @{}}
      \toprule
      \textbf{Model} & \textbf{F1} & \textbf{Prec.} & \textbf{Rec.} & \textbf{FPR} & \textbf{FNR} \\
      \midrule
      Seattle OncoID & 0.869 & 0.775 & 0.990 & 0.819 & 0.010 \\
      KY OncoID      & \textbf{0.950} & \textbf{0.907} & \textbf{0.997} & \textbf{0.290} & \textbf{0.003} \\
      \bottomrule
    \end{tabular*}
  }
\end{table}



\subsection{Systematic Manual Review of Labels}
Our dataset for manual review had an initial positive prevalence rate of 0.500 with 300 positive (reportable) and 300 negative (non-reportable) pathology reports. After three independent manual reviews of these reports, the positive prevalence rate (PPR) was estimated to be significantly lower at 0.398 (p-value=0.0004) based on a Chi-Square test, shown in Table~\ref{tab:manualreview1}. This significant difference between the original and estimated positive prevalence rates suggest that approximately 20\% of our \textit{reportable} reports are mislabeled. 

In the manual review dataset, the category assigned to the \textit{reportable} class most often was \textit{Needs Review} (n=293), followed by \textit{History Of Cancer} (n=6) and \textit{Metastases} (n=1). For the \textit{non-reportable} class, most of the reports came from the $SM_-$ dataset (n=243), with smaller numbers having a category of \textit{No cancer} (n=38) and \textit{Project-Specific} (n=19) assigned by an ODS. 

Only one pathology report changed classification from \textit{non-reportable} to \textit{reportable} after manual review. This pathology report was originally categorized as \textit{No cancer}, but two of the three reviewers identified the phrase, ``New diagnosis CML'', as a SEER reportable finding for chronic myeloid leukemia (CML). The KY OncoID model also predicted this report to be \textit{reportable} in the holdout set. Conversely, 62 of the reports originally marked as \textit{reportable} were reclassified as \textit{non-reportable} during manual review. Of these, 6 were originally categorized as \textit{History of Cancer} and 56 were \textit{Needs Review}. The KY OncoID model only predicted one of these 62 to be \textit{non-reportable}. 

\begin{table}[t]
\floatconts
  {tab:manualreview1}
  {\caption{Systematic Manual Review Results}}
  {%
    \scriptsize
    \setlength{\tabcolsep}{4pt}
    \begin{tabular*}{\linewidth}{@{\extracolsep{\fill}} l c c c @{}}
      \toprule
      & \textbf{Reportable} & \textbf{Non-Rep.} & \textbf{Pos. Prev. Rate} \\
      \midrule
      Original Label & 300 & 300 & 0.500 \\
      Reviewer 1     & 247 & 353 & 0.412 \\
      Reviewer 2     & 219 & 381 & 0.365 \\
      Final Label    & 239 & 361 & 0.398 \\
      \bottomrule
    \end{tabular*}
  }
\end{table}

\subsection{Qualitative Manual Review of Misclassified Reports}

 Our KY OncoID model made 478 false negative and 25,448 false positive predictions on the complete holdout set $H$ of 418,516 reports. During a qualitative review of 30 randomly chosen false negative reports, the authors were unable to identify any reportable findings in 26 of these reports. This lack of textual evidence for reportable findings suggests label noise where some pathology reports are incorrectly assigned the \textit{reportable} class in the holdout set. Many of these reports had definitively non-reportable diagnoses using negation language such as ``Negative for malignant cells'' and ``LYMPHOCYTES ABSENT, NO EVIDENCE OF MALIGNANCY''. In the 4 remaining false negative reports subjected to review, the authors found reportable language suggesting missed cases, but they were all rare cases that may not be well represented in the training set. These included a benign brain condition, a rare head and neck cancer, a reportable condition of the renal pelvis, and a malignant cervical condition.

 From the much larger false positive population of reports, the authors also sampled 30 reports for review, 10 each from the 3 holdout sets. Of these, 4 reports had obvious reportable language in common primary sites such as ``Invasive nonkeratinizing squamous cell carcinoma'', ``Malignant pleural effusion'' and ``Ovary positive for malignancy'', and ``RIGHT LOWER LOBE:  Malignant''. In addition, 3 reports contained reportable findings for a reportable, non-malignant endometrial condition and 2 had clear reportable findings for rare cancers. The findings in the remaining 21 false positive reports could be broadly categorized as non-reportable skin and cervical conditions,  non-reportable gastrointestinal stromal tumors, and various findings with ambiguous terminology. Considering our sample was small, finding 9 documents with \textit{non-reportable} labels in our holdout with clearly reportable language suggests ground truth label noise among the false positive reports. These qualitative findings, while not rigorous, match the findings of the systematic manual review. 

\section{Conclusions and Future Work}

This study shows that training a supervised classifier on in-domain pathology reports that match local formatting, language, and terminology improves performance. The Kentucky OncoID model reduced the false positive rate to 9.7\% and the false negative rate to 0.3\%, which should meaningfully reduce human reviewer workload and improve case capture. Beyond these performance gains, our goal is to share a repeatable data curation and training strategy that other cancer registries can adopt using their own readily available data. The approach stratifies linked and unlinked reports and uses a production-matched holdout dataset so teams can adapt models to local language while controlling false negatives and keeping false positives manageable.

Building high-quality labeled pathology datasets remains difficult because ODS categorization and linkage to SEER cases occur under tight timelines, which can introduce label noise and miss real-world complexity. Reportability definitions vary across surveillance goals, abstraction standards, and state requirements, while constantly evolving with clinical practice. Facility and physician language differences add further uncertainty. To address these challenges in future work, we will explore label smoothing that uses softened targets when reviewers disagree, and we will continue in-domain training with routine operational updates and systematic validation. We will also evaluate trustworthiness by testing robustness to facility phrasing, rare primaries, and HL7 format changes, and by checking stability across subgroups defined by demographic and clinical characteristics. These audits will guide retraining, threshold tuning, and targeted data augmentation. Differences across catchment areas, as seen between Seattle and Kentucky, merit further study, and although privacy constraints limited cross-registry analyses here, data-sharing agreements that enable collaborative qualitative comparisons would be valuable.

Another promising research direction is the use of GPT-style large language models (LLMs). Although these models have demonstrated impressive capabilities in biomedical NLP, their deployment in a cancer registry setting is hindered by strict data privacy requirements, limited GPU resources, and the lack of scalable validation frameworks to guard against hallucinations and bias. Future exploration of LLMs in this context would require either substantial investment in local computational infrastructure or the establishment of data agreements with an external service provider.

\begin{camera}
\acks{The authors are extremely grateful to Stephanie Carmack, Michele Hoskins, Lisa Witt, and Tracy Sumler of the Kentucky Cancer Registry for their work reviewing and annotating pathology reports, as well as their explanations of the nuances of cancer reportability at a SEER registry.
This research was supported by the Cancer Research Informatics Shared Resource Facility of the University of Kentucky Markey Cancer Center (P30CA177558), NCI-SEER (HHSN261201800013I / HHSN26100001) and CDC-NPCR (NU58DP007144). This work has been supported in part by the Joint Design of Advanced Computing Solutions for Cancer program established by the US Department of Energy (DOE) and the NCI of the National Institutes of Health. This work was performed under the auspices of the DOE by Argonne National Laboratory under Contract DE-AC02-06-CH11357, Lawrence Livermore National Laboratory under Contract DEAC52-07NA27344, Los Alamos National Laboratory under Contract DE-AC5206NA25396, and Oak Ridge National Laboratory under Contract DE-AC05-00OR22725.}
\end{camera}

\begin{anon}
\acks{Acknowledgments currently anonymized}
\end{anon}

\bibliography{jmlr-manuscript}

@misc{bardi,
   author = {Murdock, D.and Krawczuk P.},
   title = {BARDI},
   publisher = {USDOE},
   DOI = {doi:10.11578/dc.20240328.2.},
   url = {https://github.com/DOE-NCI-MOSSAIC/bardi},
   type = {Computer Program}
}

@Article{fresco,
   author = {Spannaus, Adam and Gounley, John and Shekar, Mayanka Chandra and Fox, Zachary R and Mohd-Yusof, Jamaludin and Schaefferkoetter, Noah and Hanson, Heidi A},
   title = {FrESCO: Framework for Exploring Scalable Computational Oncology},
   journal = {Journal of Open Source Software},
   volume = {8},
   number = {89},
   ISSN = {2475-9066},
   year = {2023},
   type = {Journal Article}
}

@misc{statecancerprofiles,
  author       = {{National Cancer Institute}},
  title        = {State Cancer Profiles},
  howpublished = {\url{https://statecancerprofiles.cancer.gov/index.html}},
  note         = {Accessed: 2025-06-19}
}

@misc{seercoding2024,
  author       = {{National Cancer Institute}},
  title        = {SEER Program Coding and Staging Manual 2024},
  howpublished = {\url{https://seer.cancer.gov/tools/codingmanuals/}},
  note         = {Accessed: 2025-05-31}
}

@article{rios2019cross,
  title={Cross-registry neural domain adaptation to extract mutational test results from pathology reports},
  author={Rios, Anthony and Durbin, Eric B and Hands, Isaac and Arnold, Susanne M and Shah, Darshil and Schwartz, Stephen M and Goulart, Bernardo HL and Kavuluru, Ramakanth},
  journal={Journal of biomedical informatics},
  volume={97},
  pages={103267},
  year={2019},
  publisher={Elsevier}
}

@article{burger2016natural,
  title={Natural language processing in pathology: a scoping review},
  author={Burger, Gerard and Abu-Hanna, Ameen and de Keizer, Nicolette and Cornet, Ronald},
  journal={Journal of clinical pathology},
  volume={69},
  number={11},
  pages={949--955},
  year={2016},
  publisher={BMJ Publishing Group}
}

@article{santos2022automatic,
  title={Automatic classification of cancer pathology reports: a systematic review},
  author={Santos, Thiago and Tariq, Amara and Gichoya, Judy Wawira and Trivedi, Hari and Banerjee, Imon},
  journal={Journal of Pathology Informatics},
  volume={13},
  pages={100003},
  year={2022},
  publisher={Elsevier}
}

@inproceedings{yoon2018filter,
  title={Filter pruning of convolutional neural networks for text classification: a case study of cancer pathology report comprehension},
  author={Yoon, Hong-Jun and Robinson, Sarah and Christian, J Blair and Qiu, John X and Tourassi, Georgia D},
  booktitle={2018 IEEE EMBS International Conference on Biomedical \& Health Informatics (BHI)},
  pages={345--348},
  year={2018},
  organization={IEEE}
}

@inproceedings{martinez2011information,
  title={Information extraction from pathology reports in a hospital setting},
  author={Martinez, David and Li, Yue},
  booktitle={Proceedings of the 20th ACM international conference on Information and knowledge management},
  pages={1877--1882},
  year={2011}
}

@article{schroeck2017development,
  title={Development of a natural language processing engine to generate bladder cancer pathology data for health services research},
  author={Schroeck, Florian R and Patterson, Olga V and Alba, Patrick R and Pattison, Erik A and Seigne, John D and DuVall, Scott L and Robertson, Douglas J and Sirovich, Brenda and Goodney, Philip P},
  journal={Urology},
  volume={110},
  pages={84--91},
  year={2017},
  publisher={Elsevier}
}

@article{saib2020hierarchical,
  title={Hierarchical deep learning classification of unstructured pathology reports to automate ICD-O morphology grading},
  author={Saib, Waheeda and Chiwewe, Tapiwa and Singh, Elvira},
  journal={arXiv preprint arXiv:2009.00542},
  year={2020}
}

@article{khosravi2021deep,
  title={A deep learning approach to diagnostic classification of prostate cancer using pathology--radiology fusion},
  author={Khosravi, Pegah and Lysandrou, Maria and Eljalby, Mahmoud and Li, Qianzi and Kazemi, Ehsan and Zisimopoulos, Pantelis and Sigaras, Alexandros and Brendel, Matthew and Barnes, Josue and Ricketts, Camir and others},
  journal={Journal of Magnetic Resonance Imaging},
  volume={54},
  number={2},
  pages={462--471},
  year={2021},
  publisher={Wiley Online Library}
}

@article{gao2019classifying,
  title={Classifying cancer pathology reports with hierarchical self-attention networks},
  author={Gao, Shang and Qiu, John X and Alawad, Mohammed and Hinkle, Jacob D and Schaefferkoetter, Noah and Yoon, Hong-Jun and Christian, Blair and Fearn, Paul A and Penberthy, Lynne and Wu, Xiao-Cheng and others},
  journal={Artificial intelligence in medicine},
  volume={101},
  pages={101726},
  year={2019},
  publisher={Elsevier}
}

@article{hsu2024machine,
  title={Machine learning and deep learning tools for the automated capture of cancer surveillance data},
  author={Hsu, Elizabeth and Hanson, Heidi and Coyle, Linda and Stevens, Jennifer and Tourassi, Georgia and Penberthy, Lynne},
  journal={JNCI Monographs},
  volume={2024},
  number={65},
  pages={145--151},
  year={2024},
  publisher={Oxford University Press}
}

@book{fleiss2013statistical,
  title={Statistical methods for rates and proportions},
  author={Fleiss, Joseph L and Levin, Bruce and Paik, Myunghee Cho},
  year={2013},
  publisher={john wiley \& sons}
}

\end{document}